# Utilizing Smartphone-Based Machine Learning in Medical Monitor Data Collection: Seven Segment Digit Recognition

Varun N. Shenoy[1], Oliver O. Aalami, M.D.[2]
[1]Cupertino High School, Cupertino, California, [2]Stanford University, Palo Alto, California

**Abstract**

*Biometric measurements captured from medical devices, such as blood pressure gauges, glucose monitors, and weighing scales, are essential to tracking a patient's health. Trends in these measurements can accurately track diabetes, cardiovascular issues, and assist medication management for patients. Currently, patients record their results and date of measurement in a physical notebook. It may be weeks before a doctor sees a patient's records and can assess the health of the patient. With a predicted 6.8 billion smartphones in the world by 2022[1], health monitoring platforms, such as Apple's HealthKit[2], can be leveraged to provide the right care at the right time. This research presents a mobile application that enables users to capture medical monitor data and send it to their doctor swiftly. A key contribution of this paper is a robust engine that can recognize digits from medical monitors with an accuracy of 98.2%.*

**Introduction**

The goal of this research is to develop and evaluate the accuracy of a smartphone based system to automatically recognize and record biometric monitor results to Apple's HealthKit (Apple Inc., Cupertino, Ca). As the internet becomes more embedded into medical monitors through Wi-Fi and Bluetooth technologies, people who cannot afford to upgrade to these expensive machines will fail to receive the benefits of rapid medical attention. Our solution relies on a common smartphone to do the data capture, processing and communication. We use computer vision based feature extraction alongside machine learning to decipher a reading of a medical monitor captured by a smartphone. The medical monitors generally use seven-segment displays (where each digit is made up of a combination of segments). The model proposed in this paper can accurately read the monitor 98.2% of the time.

**Prior Work**

Whilst seven-segment digit recognition technology has been created in the past, prior work has not accounted for conditions affected by variable lighting and positioning. One proposed solution was ssocr[3], an acronym for Seven Segment Optical Character Recognition. Such an approach was unsuitable to actively adapt to dynamic orientations and lighting of a smartphone image. Tesseract[4], Google's popular optical character recognition model, was also another choice. However, we learned that while Tesseract is strong at reading regular text on a page, it had a difficult time accurately reading seven segment displays. Prior work in the field of medical informatics focussed on the recognition of hand written characters[5].

**Implementation**

We delineate implementation into three phases. The first part will cover the use model of the app, the second part will cover the front end implementation and user interface. Finally, we will discuss the backend digit recognition approach.

*Application Use Model*

The use model is described as follows. Consider the case of measuring a biomarker such as blood glucose. After the glucose monitor displays a reading, the patient opens the smartphone application and selects the "Glucose" option. Next, the patient scales a bounding rectangle to fit the digits on the display. These bounds are saved for the next time a measurement of glucose will be taken. After taking a picture, the image within the bounds is sent to a backend server using wireless communications via the internet. A trained machine learning Random Forest[6] classifier with image processing based feature extraction divides the image into individual digits and classifies each of them. These numbers are sent back to the user's device. The user is presented with the server's prediction for the digits in the image. The user can easily correct the predicted numbers by using a scroll picker for each of the digits. This data is then saved into Apple HealthKit and is readily available for the physician.

*Front-end Implementation and User Interface*

The front end implementation of the application is focused on an easy and intuitive use model. The target audience is mainly the elderly who lack the dexterity required to accurately use mobile keyboards. As such, it is imperative that the font sizes throughout the mobile application are large and bold to aid the user's eyes in reading. We also implemented a voice assistant to guide the user through the steps and in reading the result. Ultimately, the goal was to make the final user flow (Figure 1) of the app succinct and user-friendly. Figure 1 shows the steps involved in measuring systolic blood pressure through our application.

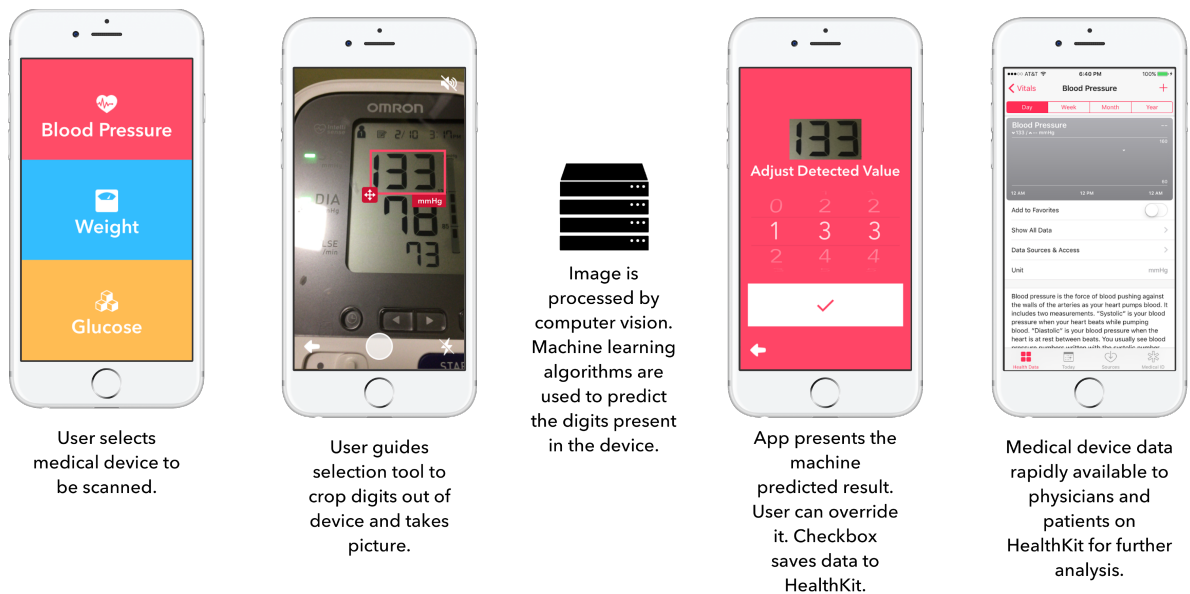

**Figure 1.** The User Interface Flow for the Mobile Application

On the front end side of the application, the Swift 3.0.x[7] programming language was used for developing the iOS app and interfacing with HealthKit in the Xcode 8.3.x[8] integrated development environment. Alamofire[9], a networking framework, was used to simplify the process of sending the image from the phone to the server via Hyper Text Transfer Protocol (HTTP) in a multipart form. Since our model was not embedded into the device, and lives on a server, computational requirements for the client are minimal. Our application will work as long as the device can make HTTP calls and access the internet. Apple's AVSpeechSynthesizer class was used in the app to provide an active voice assistant in the app. The voice tells the patient exactly how to take a picture and reads out the resulting digits when it is saved to HealthKit. All components of the app were optimized to work on any Apple iPhone running iOS 8 and above.

*Backend Digit Recognition Approach*

A basic block diagram (Figure 2) shows the backend methodology for the server side digit recognition algorithm. The model is first trained on a training set of images and then tested on a new set of images it has never seen before.

We chose to implement our algorithmic model in Python 2.7.x[10], since there exist a diverse set of libraries for machine learning and computer vision. We used the scikit-learn[11] package to implement a Random Forest classifier and model persistence. OpenCV[12] for Python was used for feature extraction and image normalization due to its versatility and high computation with a simple programming interface. The Matplotlib[13] graphics library was used to create data visualizations for this research.

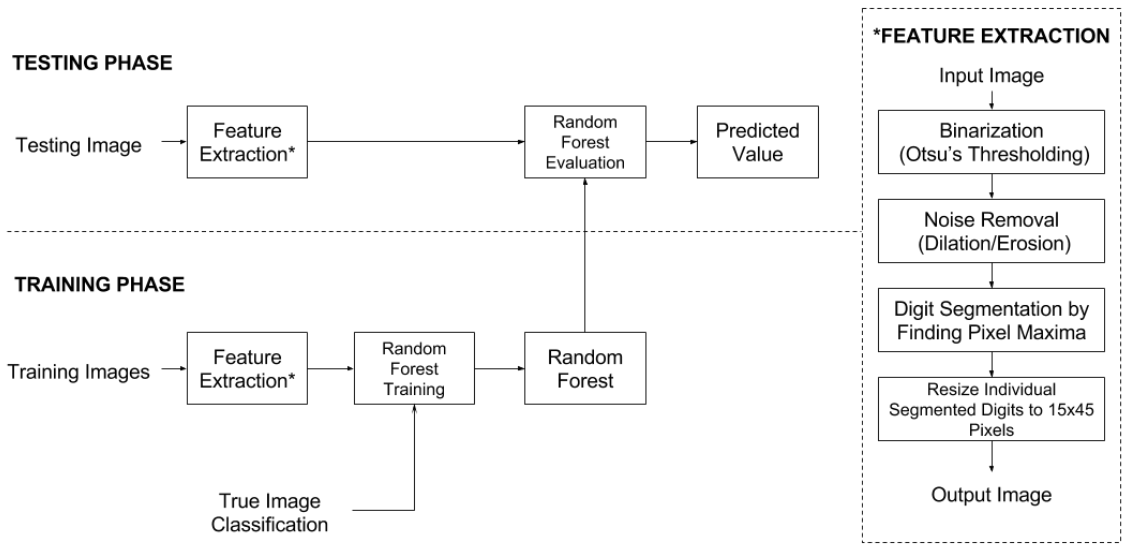

**Figure 2.** Image Processing and Machine Learning Methodology.

A shared aspect of the training and testing steps is the ability to separate an image into disparate digit components. This is called feature extraction. The input image is divided into individual digits so the classifier only has to work with ten possible classes (0-9). Image processing is used to reduce the complexity of input data and normalize the images for the machine learning model. OpenCV's built-in Otsu's thresholding algorithm[14] is used to reduce the given image to just black and white pixels. Otsu's method dynamically separates the foreground and background of the input image and maximizes the variance between the two. After thresholding, we segment the image into individual digits using a vertical projection algorithm. A graph for the average pixel color in each column of the image is created (Figure 3). Peaks on the graph where the average pixel value is white (255) should be disregarded and gives us a location for digit segmentation. Each column is scanned and a splice is created wherever the average color is close to white. All sections of pure white are deleted and each individual digit is separated to be classified in the machine learning model. All of the digits are scaled to 15x45 pixels to minimize variance between images with the same digit in them.

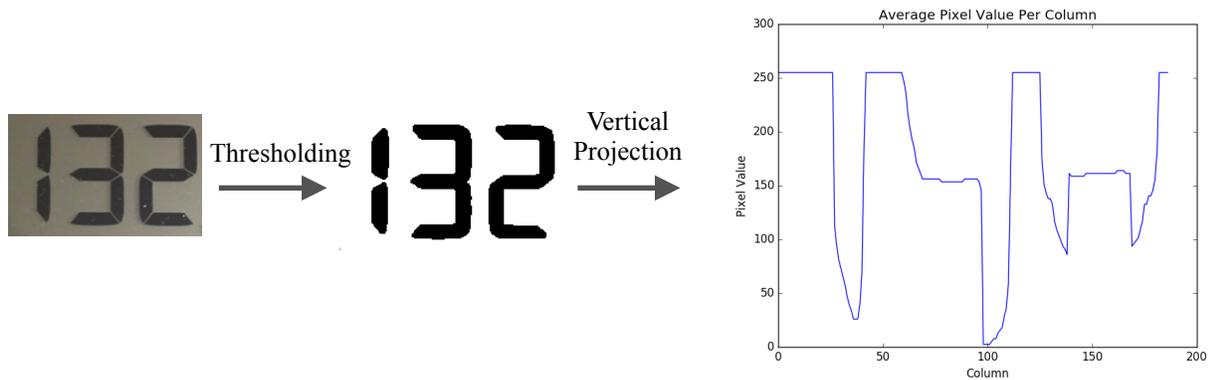

**Figure 3.** Feature Extraction Process on an Example Image.

A Random Forest is a machine learning classifier that is composed of an ensemble of decision tree estimators. Each decision tree is a tree-like model of decisions and their possible consequences, ultimately leading to a final conclusion (leaf node). While decision trees on their own are weak classifiers, Random Forests are fast, accurate, and not prone to overfitting. In this research, 100 trees were used in the forest for rapid training and testing. Trees in the forest are grown using bootstrap aggregating, or bagging. Each tree is constructed on a random subset of the input data, so they produce different models that can be averaged. This ensures that the model cannot be overfitted by the training data. When a new, unlabeled, piece of data is shown to the random forest, each tree assesses it individually. The forest makes a decision through majority voting by the trees. The Random Forest classifier machine learning model was chosen over others, since it is very fast and accurate. The fully trained Random Forest ensemble for this research only takes up 847 kB of space and takes under 3 seconds to train on a 2.9GHz processor for laptops. Calls to the server take 2-3 seconds to process and report data back to the client device.

**Experimental Results**

The past records of a patient's heart pressure monitor and a weighing scale were used to cultivate training and testing data (Figure 4). We sampled the historical data stored within these devices to recover 108 images for the testing set. These images had 1-3 seven segment digits, which our algorithm had to segment and classify. The software was trained on a small additional subset of representative seven segment displays (25 images). Screenshots of digits typed using a seven-segment font were used as training data as well. Images in the training set were not used in the testing set.

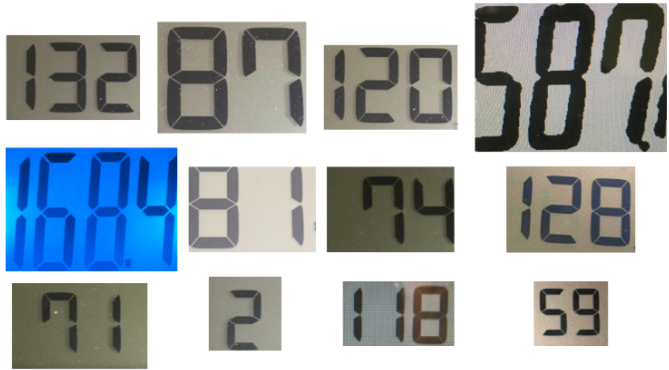

**Figure 4.** Example Images Used to Train and Test the Model.

Furthermore, we observed a non-uniform distribution of digits in the collected samples (Figure 5). E.g. there were 85 images of "1"s, significantly more than the rest. This was expected as systolic blood pressure values in healthy adults have a leading "1". The same can be said for weight measurements as leading "1"s and "2"s are the most common.

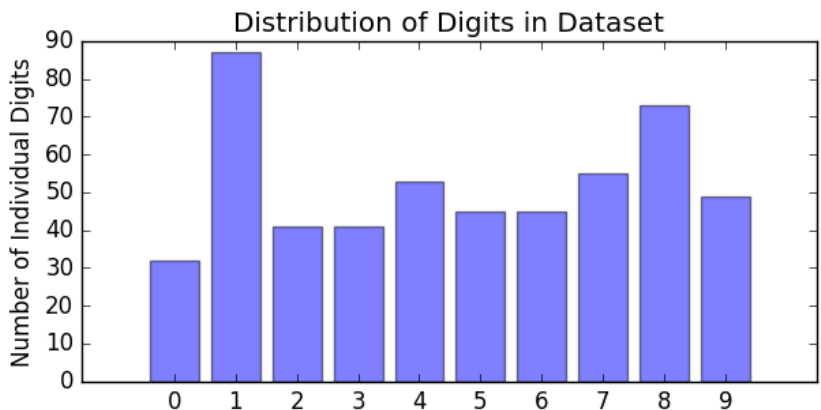

**Figure 5.** The Distribution of Digits throughout the Entire Dataset.

The machine learning model successfully recognized all of the digits except in two cases, achieving an accuracy of 98.2%. We examined three dominant features using Principal Component Analysis[15] (PCA) on the testing and training set. This helps us observe the strength of our algorithm by visualizing clustering patterns determined by the 3-component PCA (Figure 6). In these three-dimensional plots, digits of the same class are observed to be well clustered. However, certain digits are clumped closer together than others. For example, the digits "0" and "8" are clustered together. This means that visually, digits that are "8"s and "0"s are similar, and therefore, we require more data so the model can learn concrete differences between the two. Also, since we reduced every vector to three dimensions from 675, certain information that can be crucial to differentiating "8"s and "0"s may have been lost.

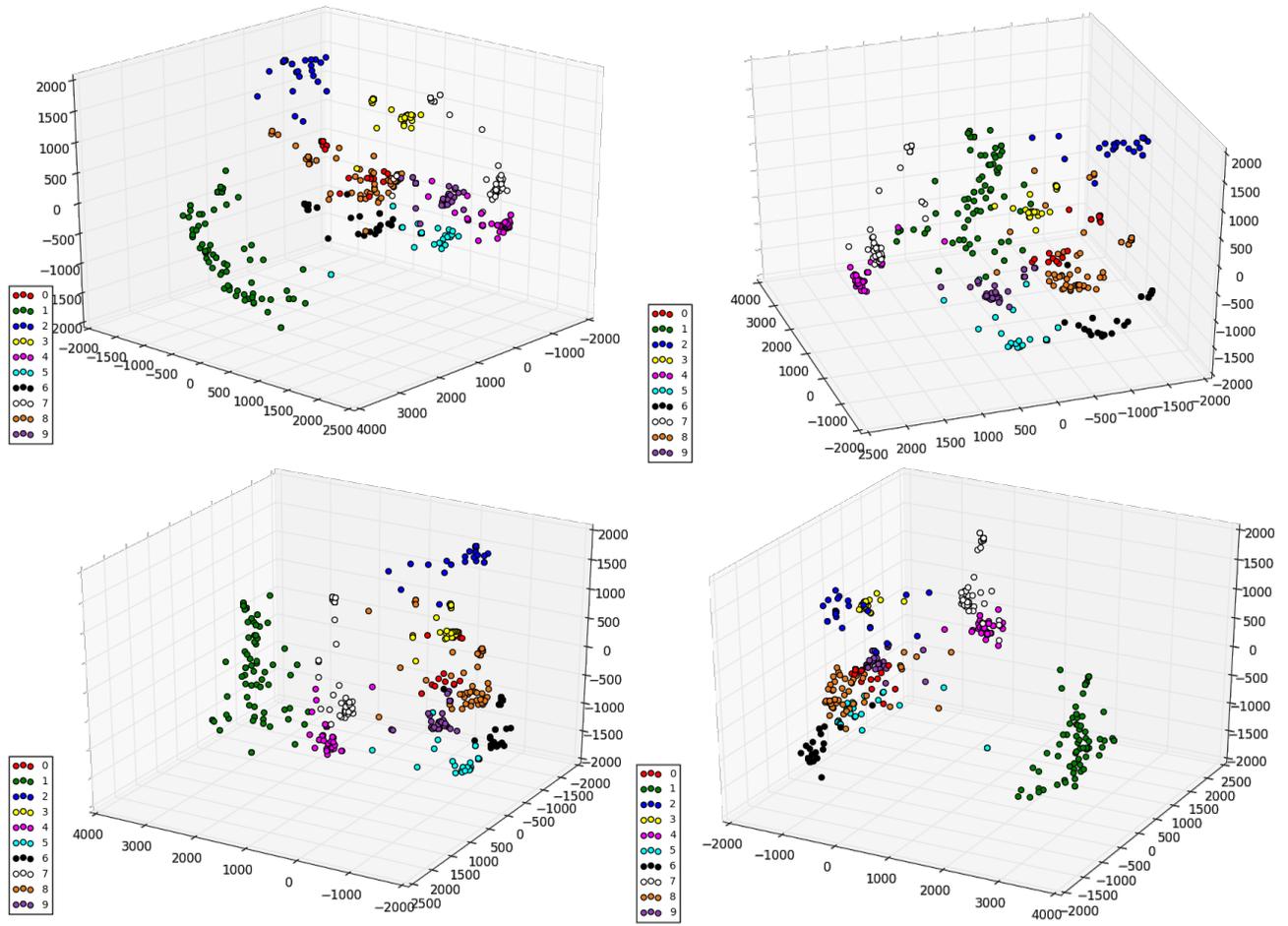

**Figure 6.** Different Orientations of PCA-Reduced Data in Three-Dimensions.

We analyzed the output of our model on our testing data and we found that it performed well across all 10 classes (Figure 7). Across the 262 individual digits in the testing set (post image segmentation), we achieved an overall accuracy of 98.2% and an F1-score of 0.978 with our Random Forest model. An F1-score is a metric for classification which is best at 1 and worst at 0. It is the harmonic mean of precision and recall.

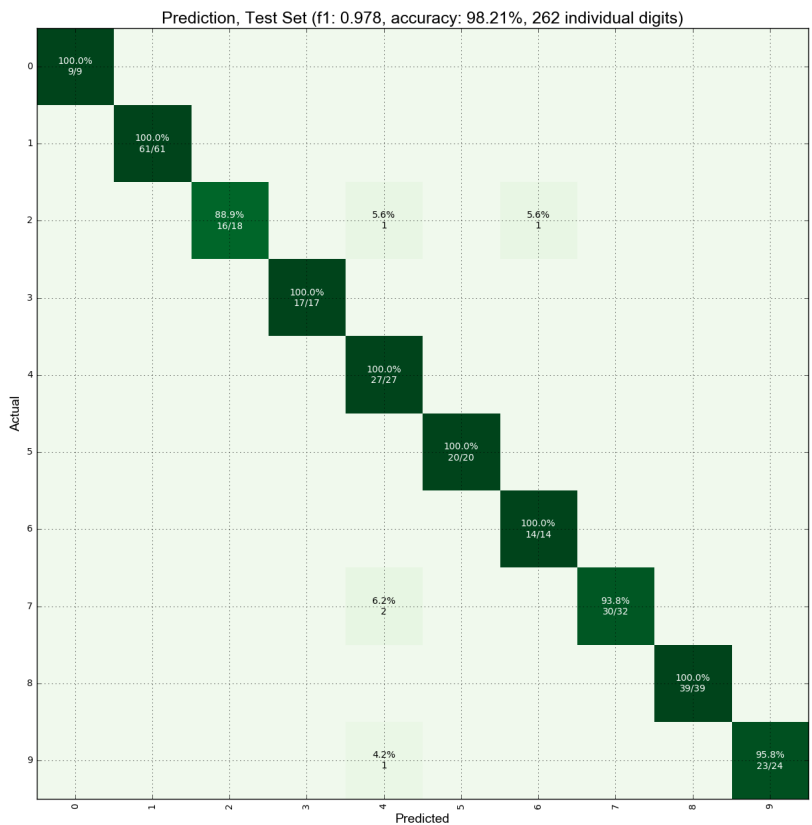

**Figure 7.** Confusion Matrix of Our Model on All 10 Classes.

Other machine learning approaches from the scikit-learn library were used as well to verify the strength of our proposed model (Table 1). Ultimately, the Random Forest model achieved the highest accuracy of 98.2% while the General Purpose Multi-layer Perceptron was only accurate 61.1% of the time. Furthermore, a single decision tree (a voting member of a random forest model trained on the entire training set) attained an accuracy of 64.8%. Random Forests did the best since it is very difficult to overfit on the testing data. Other models are more prone to overfitting on the training data and as such, fail to recognize testing data correctly.

**Table 1.** Other Tested Machine Learning Models and Their Respective Accuracies from scikit-learn

| **Classifier** | **Accuracy** |
| --- | --- |
| Random Forest | 98.2% |
| Linear Support Vector Machine | 94.4% |
| Decision Tree | 64.8% |
| Naive Bayes | 86.1% |
| K-Nearest Neighbors (neighbors = 5) | 92.6% |
| General Purpose Multi-layer Perceptron | 61.1% |

**Conclusion**

In this research, we presented the design and implementation of a smartphone application that can capture medical data and communicate it to physicians. Using our tool, physicians can now catch early trends and diagnose critical issues before the onset of the medical vulnerability. We described a versatile seven segment digit recognition algorithm that can be used to rapidly read any digital display via computer vision and machine learning. Special attention was given to the user interface to serve the ill and elderly.

Since the submission of the paper for review, we have extended our backend infrastructure by uploading it to a cloud server (Amazon Web Services Lambda).

**Further Work**

There are many ways to further improve our algorithm. For one, sharp images guarantee better results. We would like to consider blur detection prior to analyzing our image. If the image is too blurry, we can send a message back to the user, requesting a clearer picture. There are many well known techniques to accurately measure blur within an image.

On a larger scale, it is necessary to gather more images for both training and testing. Creating a robust corpus of images will enable us to improve the accuracy of our method. We plan to test the software with real patients to validate the user interface in a medical environment.

We would also like to look into embedding our model into mobile devices directly without the need for a server. This will drastically increase speed for users and enable them to use the app in locations without access to the internet. State-of-the-art research is currently focussed on mobile embedded machine learning[16].